%% file: main.tex
\documentclass[letterpaper, 10 pt, conference]{ieeeconf}  

\IEEEoverridecommandlockouts                              

\usepackage{graphics} 
\usepackage{multirow}
\usepackage{epsfig} 
\usepackage{times} 
\usepackage[caption=false,font=footnotesize]{subfig}
\usepackage{fixltx2e}
\usepackage{amsmath} 
\usepackage{amssymb}  
\usepackage{url}
\usepackage[dvipsnames]{xcolor}

\title{\LARGE \bf
Is it Worth to Reason about Uncertainty in Occupancy Grid Maps during Path Planning?
}

\author{Jacopo Banfi$^{1*}$, Lindsey Woo$^{2*}$, and Mark Campbell$^{2}$%
\thanks{$^{*}$Equal contribution.}
\thanks{$^1$CSAIL, Massachusetts Institute of Technology (MIT), Cambridge (MA) 02139, USA. Email: jbanfi@mit.edu. Work done while at Cornell University.}
\thanks{$^2$Sibley School of Mechanical and Aerospace Engineering, Cornell University, Ithaca (NY) 14853, USA. Email: \{lw734,mc288\}@cornell.edu.}
\thanks{This work was supported by the MURI grant N00014-17-1-2699.}}

\begin{document}

\maketitle

\input{content/abstract}

\input{content/introduction}
\input{content/related_work}
\input{content/problem}
\input{content/method}
\input{content/experiments}

\input{content/discussion}




\addtolength{\textheight}{-14.5cm}   










\bibliographystyle{IEEEtran}
\bibliography{IEEEabrv, references}

\end{document}

%% file: content/abstract.tex
\begin{abstract}
This paper investigates the usefulness of reasoning about the uncertain presence of obstacles during path planning, which typically stems from the usage of probabilistic occupancy grid maps for representing the environment when mapping via a noisy sensor like a stereo camera. The traditional planning paradigm prescribes using a hard threshold on the occupancy probability to declare that a cell is an obstacle, and to plan a single path accordingly while treating unknown space as free. We compare this approach against a new uncertainty-aware planner, which plans two different path hypotheses and then merges their initial trajectory segments into a single one ending in a ``next-best view'' pose. After this informative view is taken, the planner commits to one of the hypotheses, or to a completely new one if a collision is imminent. Simulations were conducted comparing the proposed and traditional planner. Results show the existence of planning scenarios ---like when the environment contains a dead-end, or when the goal is placed close to an obstacle--- in which reasoning about uncertainty can significantly decrease the robot's traveled distance and increase the chances of reaching the goal. The new planner was also validated on a real Clearpath Jackal robot equipped with a ZED 2 stereo camera.

\end{abstract}

%% file: content/introduction.tex
\section{INTRODUCTION}

Navigation in unstructured environments is challenging for an autonomous mobile robot due to the requirement of jointly mapping, localization and planning ---all in real time. This is especially true when the environment is very cluttered (e.g.\ a disaster recovery scenario) and the robot can only rely on noisy sensors (e.g.\ cameras detecting obstacles with potential occlusions). For example, stereo cameras provide several advantages over lidar (lighter, cheaper, and less power-hungry), but these come at the expense of a rapid degradation in the quality of depth information~\cite{wang2019pseudo}. Probabilistic occupancy grids, like those produced by the Octomap mapping framework~\cite{hornung13auro}, provide a principled way of integrating noisy measurements into a 3D map that can be used for navigation. However, as of today, the mainstream approach to online path planning over grid maps still neglects the presence of uncertainty in the cells' occupancy, treating them as either free, occupied, or unknown, and plans a single low-cost path to the goal assuming that unknown cells are free~\cite{ryll2019efficient,tordesillas2020faster}. The presence of unexpected obstacles is simply dealt with by replanning a new path. While using a hard threshold on occupancy probability provides an easy way to label a cell as ``obstacle'' (e.g.\ as used by the Octomap ROS package\footnote{\url{http://wiki.ros.org/octomap_ros}}), this only works well  if depth measurements are reliable, which might only be true for short ranges.

\begin{figure}[t!]
    \centering
    \includegraphics[width=\columnwidth]{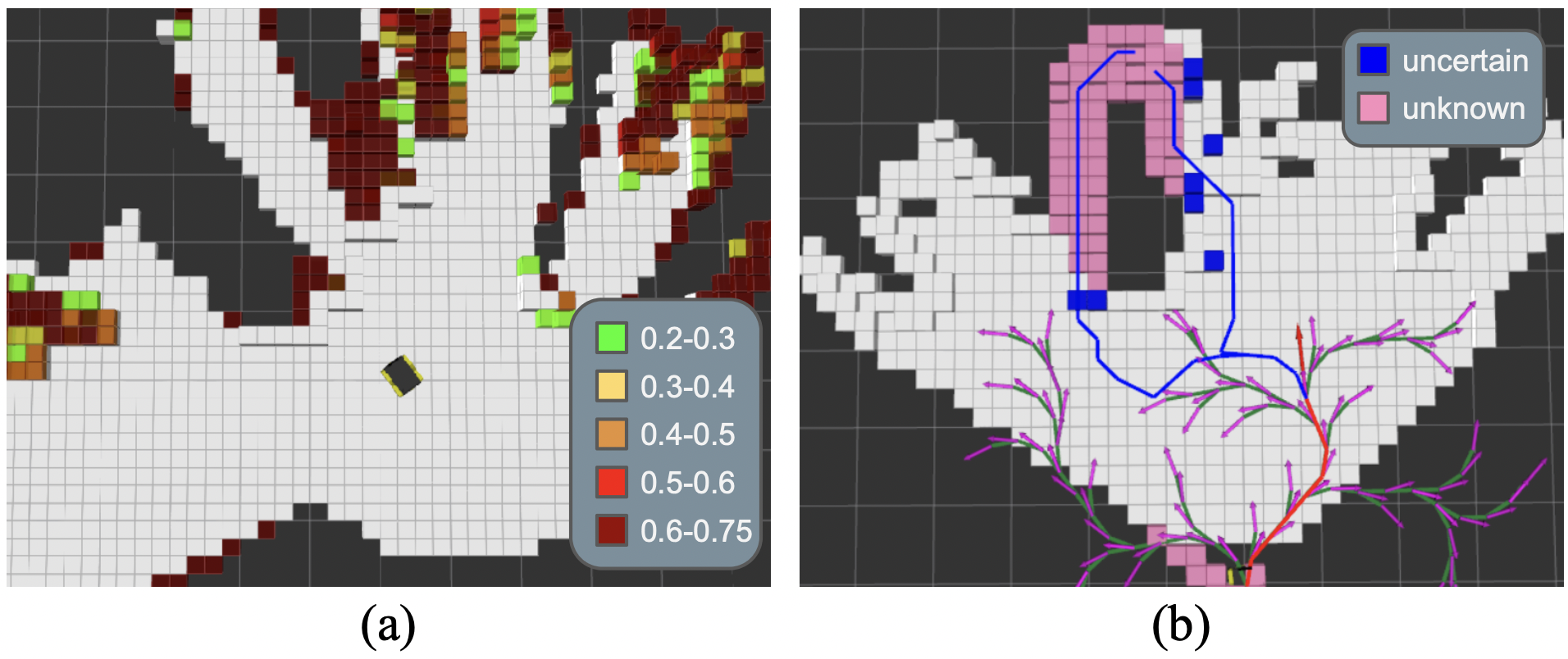}
    \vspace*{-7mm}
    \caption{The NBV planner in action in a simulated environment. (a): Example Octomap map. White cells: ground and obstacles cells with occupancy probability $\geq$ 0.75; colored cells: uncertain occupancy probability, but potentially safe. (b): NBV planner first stage. Candidate NBV poses are in purple, the best one is in red (along with the first common trajectory segment) and the two path hypotheses are in blue.}
    \label{fig:qual}
    \vspace{-4mm}
\end{figure}

This work presents a novel approach to directly reason about the presence of uncertain obstacles in occupancy grid maps during the global path planning phase. Instead of planning a single path on a deterministic, optimistic map, our planner computes different path hypotheses and then merges their initial trajectory segments into a single one ending in a ``next-best view'' (NBV) pose. This is obtained from a set of candidate poses by maximizing an objective function combining entropy reduction, probability of occlusion, and proximity to the cells lying along the planned paths; see Fig.~\ref{fig:qual}. After this informative view is taken, the planner commits to one of the hypotheses, or to a completely new one if a collision is imminent. 

We study the performance of the proposed next-best view planner against the traditional approach in extensive simulations. Our results show the existence of scenarios ---like when the environment contains a dead end, or when the goal is placed close to an obstacle--- in which reasoning about uncertainty can significantly decrease the robot's traveled distance and increase the chances of reaching the goal, especially when high noise is present. A validation of the planner on a real Clearpath Jackal robot equipped with a ZED 2 stereo camera complements the simulation results.\footnote{The accompanying video shows example runs of simulations and real world experiments.} 


%% file: content/related_work.tex
\section{RELATED WORK}

When one thinks about cost-optimal global path planning for a mobile robot, some version of A* coupled with the usage of a binary occupancy grid is the pair which immediately comes to mind. Perhaps surprisingly, not much has changed since the introduction of A* in the late sixties~\cite{hart1968formal}. The main innovations were essentially three: a more efficient way of handling the replanning phase under the optimistic assumption of treating unknown space as free (with D*~\cite{stentz1994} and its variants~\cite{stentz1995, koenig2002d}), the possibility of incorporating kinematic and dynamic constraint during the global planning phase (with sampling-based algorithms like Rapidly Exploring Random Trees (RRTs)~\cite{rrt} and probabilistic roadmaps~\cite{kavraki1996probabilistic}, or with hybrid versions of A*~\cite{dolgov2008practical}), and other improvements focused on runtime (e.g. JPS~\cite{harabor2011online}) and path smoothness (e.g. Theta*~\cite{daniel2010theta}). As of today, even when considering 3-D grid maps, an approach based on A* coupled with a binary occupancy grid is still the go-to choice for what concerns global path planning even when the robot is equipped with a depth camera; see, for example,~\cite{ryll2019efficient, tordesillas2020faster}. We deem that the reasons are essentially two. First, A* is simple to understand and easy to implement. Second, replanning is a simple way of dealing with discrepancies between expected and actual obstacles. However, as our experiments show (see Section~\ref{sec:simulations}), it might be very easy to run into planning instances where this approach is suboptimal, and reasoning about map uncertainty can be beneficial. 

Map uncertainty stems not only from cells having received noisy measurements, but also from unknown regions of the environment which have not been mapped due to occlusions or limited sensing range. Spurred by the successes of deep learning, some recent works started challenging the idea of optimistically treating unknown space as free~\cite{stein2018learning,yutao2}. Unfortunately, these methods are designed to work well only when coupled with a precise depth sensor, i.e. a lidar. The multi-hypothesis planning framework of Han et al.~\cite{dshppc2020}, instead, is able to reason about uncertain terrain semantics on a previously mapped environment using only a depth camera, by using a Bayesian Neural Network to obtain a probabilistic measure of path safety. This paper is complementary to this latter work, as it offers a solution to the global planning problem in previously unmapped environments containing several physical obstacles to avoid, but borrows from it (and other works such as~\cite{hardy,tordesillas2020faster}) the idea of planning more than a single path to cope with uncertainty in the environment.

%% file: content/problem.tex
\section{PROBLEM SETTING}
\label{sec:problem}

Consider a mobile ground robot\footnote{This approach could also be generalized to an aerial vehicle.} equipped with a noisy mapping sensor. We assume that the robot can localize itself in the environment with reasonable accuracy, and to build a probabilistic occupancy grid (2-D or 3-D) as it moves around by fusing the measurements from its mapping sensor. The set of map cells is denoted by $\mathcal{M}$, which is partitioned into four sets: 
\begin{itemize}
\item $\mathcal{U}$, the set of cells belonging to unknown space;
\item $\mathcal{O}$, the set of cells denoting the presence of an obstacle;
\item $\mathcal{F}$, the set of cells associated with free space;
\item $\mathcal{N}$, the set of cells associated with uncertain occupancy.
\end{itemize}

In the above categorization, $\mathcal{U}$ denotes cells not yet associated with at least one measurement. The remaining sets can be obtained by defining two probability thresholds, $p_l$ and $p_h$, with $p_l < p_h$. Given a generic cell $c \in \mathcal{M} \setminus \mathcal{U}$, $p(c)$  denotes the corresponding occupancy probability. If $p(c) \geq p_h$, then $c \in \mathcal{O}$; if  $p(c) \leq p_l$, then $c \in \mathcal{F}$; otherwise,  $c \in \mathcal{N}$.

In this paper, we consider an online path planning problem in which the robot drives from an initial start pose $\mathbf{p}_s$ to a given goal region $X_{\text{goal}}$ while avoiding collisions with obstacles and traveling as little distance as possible. {\em We assume that the goal region lies either within the robot's mapping range, or at least in its proximity.} Thus, the goal region could be thought as a ``local subgoal'' provided by a higher-level path planning module.

A path is represented as a sequence of $(x,y,\theta)$ tuples in which the orientation might be left unspecified. In that case, we assume that a local planner would still be able to track the planned path with reasonable accuracy. We use $\pi$ to denote a generic path, which has the form $$\pi= [\mathbf{p}_1=\mathbf{p}_s, \mathbf{p}_2,\ldots,\mathbf{p}_g],$$ where $\mathbf{p}_g$ denotes the goal pose which must fall inside $X_{\text{goal}}$. Given a path $\pi$, each pose $\mathbf{p}_i$ is associated with a set of cells lying within the corresponding neighborhood. We use $c(\mathbf{p}_i)$ to denote such cells. The neighborhood shape and size are chosen in order to fully contain the robot's footprint.

 \begin{figure*}[t!]
     \centering
     \includegraphics[width=0.9\textwidth]{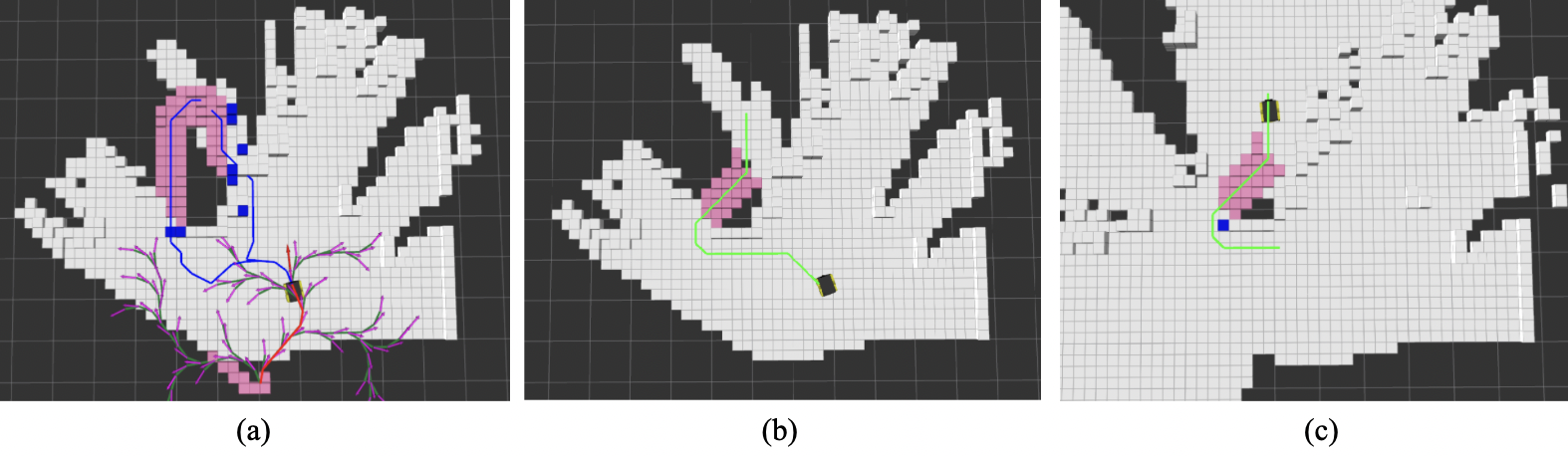}
     \vspace*{-4mm}
     \caption{Example NBV planner run. (a): NBV pose is reached; (b): The right hypothesis is discarded as it leads to a collision, while the left hypothesis is replaced with one travelling through less unknown space; (c): The robot reaches the goal after a replan in proximity of an obstacle.}
     \label{fig:example}
     \vspace*{-4mm}
 \end{figure*}

Due to the fact that the map may, in general, change as the robot moves, the robot might discover over time that the current path will result in a collision, or is longer than it could be.  Therefore, the path can be replanned as new information is acquired (e.g.\ at a fixed frequency).

A traditional approach to the path planning problem described above is place all the cells with uncertain occupancy $\mathcal{N}$ inside either the free set $\mathcal{F}$ or the occupied set $\mathcal{O}$, and plan a single path under the optimistic assumption that all the unknown cells in $\mathcal{U}$ belong to $\mathcal{F}$. The next section describes an alternative method based on a technique borrowed from the literature on 3-D object reconstruction~\cite{daudelin2017adaptable}: entropy reduction via ``next-best views.''


%% file: content/method.tex
\section{NEXT-BEST VIEW PATH PLANNING}
\label{sec:nbv}

The proposed NBV planner works in three stages. During the first stage, up to two path hypotheses can be computed. If two hypotheses are obtained, these will have a short initial trajectory segment in common ending in an uncertainty-reducing NBV pose. In the second stage, which is only executed if two hypotheses and associated NBV pose are available, the robot follows the initial trajectory segment, reaches the NBV pose, and selects a specific plan hypotheses (from the original two, and an additional hypothesis computed directly at the NBV pose). In the third and final stage, the robot follows the best hypothesis, replanning as needed. These three stages are shown in Figs.~\ref{fig:qual}-\ref{fig:example} and described in more detail below.

\subsection{Stage 1: Hypotheses and NBV Generation}
 In this initial stage, up to two path hypotheses $\mathcal{H} = \{\pi_1, \pi_2\}$ are first generated between the start pose $\mathbf{p}_s$ and the goal region $X_{\text{goal}}$.\footnote{This approach can be generalized to more than two hypotheses, which may be useful for other scenarios such as larger environments.} 
 For each of the hypotheses, all poses $\mathbf{p}_i$ should not have a cell in the corresponding neighborhood $c(\mathbf{p}_i)$  lying in $\mathcal{O}$. However, they can represent attempts to traversing cells lying in $\mathcal{U}$, $\mathcal{N}$, or both. In our current approach, orientations are not considered at this stage, and classical A* is used for planning the hypotheses. The process works as follows. The first hypothesis is planned assuming that cells lying in $\mathcal{U}$ denote free space. For what concerns cells in $\mathcal{N}$, a pose is considered not colliding if it contains at most a given number $C_\mathcal{N}$ of this type of cells in its neighborhood $c(\mathbf{p}_i)$. If the first hypothesis is not traversing any cells in $\mathcal{U}$ or $\mathcal{N}$, there is no need to compute a second one and the planner directly proceeds to Stage 3. Otherwise, we make a copy of the original probabilistic map, and mark cells lying within a given (Manhattan) distance $D_{\text{HYP}}$ to cells in $\mathcal{U}$ and $\mathcal{N}$ lying along the computed path as obstacles, unless they are close to the start and goal pose. This map is used to compute the second hypothesis. If no second hypothesis is found, the planner proceeds directly to Stage 3.
 
In case two hypotheses are obtained, we build a set of candidate NBV poses $\mathcal{P}$ by growing an RRT within a given radius $R_{\text{RRT}}$ from the current robot's pose by using a predefined set of kinematically constrained motion primitives. Fig.~\ref{fig:qual} shows the RRT (in green) along with candidate NBV poses (in purple). The RRT should be grown by making sure that all poses along each path can be safely reached. The quality of a generic NBV pose $\mathbf{p}_{\text{NBV}}$ is evaluated via an objective function $J: \mathcal{P} \rightarrow [0, 1]$ combining an entropy reduction term and a distance-related term: 
\begin{equation}
    J(\mathbf{p}_{\text{NBV}}) = \frac{\alpha}{\eta_{H}} J_{H}(\mathbf{p}_{\text{NBV}}) + \frac{(1 - \alpha)}{\eta_{d}} J_{d}(\mathbf{p}_{\text{NBV}}),
\end{equation}
where $\alpha \in [0,1]$ is a parameter that can be used to tune the importance of the two terms, and $\eta_{H}$ and $\eta_{d}$ are normalization terms computed for each term independently. 

With a slight abuse of notation, let $\mathcal{C}(\mathbf{p}_{\text{NBV}})$ represent the cells belonging to either $\mathcal{U}$ or $\mathcal{N}$ such that:
\begin{itemize}
    \item they are associated with at least one potential traversal along a path hypothesis, i.e. they belong to the set $c(\mathbf{p}_i)$ for some pose $\mathbf{p}_i$ of at least one path $\pi_{1,2}$, and
    \item the projection of their centroid falls within the camera plane assuming that the robot is located at $\mathbf{p}_{\text{NBV}}$.
\end{itemize}
 The entropy $H(c)$ of a cell $c \in \mathcal{M}$ is given by
 \begin{equation}
     H(c) = - p(c) \ln{p(c)} - (1 - p(c)) \ln (1 - p(c)),
 \end{equation} 
 with $p(c) = 0.5$ if $c \in \mathcal{U}$. The probability $V(c, \mathbf{p}_{\text{NBV}})$ that a cell $c \in \mathcal{C}(\mathbf{p}_{\text{NBV}})$ can actually be seen from pose $\mathbf{p}_{\text{NBV}}$ is computed as 
 \begin{equation}
     V(c, \mathbf{p}_{\text{NBV}})= \prod_{l=1}^{n} (1 - p(c_l)),
 \end{equation}
 \noindent where $c_l, l=1,\ldots,n$ are the $n$ cells traversed by the ray casted from the candidate viewpoint to cell $c$. 
 The entropy term $J_{H}(\mathbf{p}_{\text{NBV}})$ is computed as 
 \begin{equation}
     J_{H}(\mathbf{p}_{\text{NBV}}) = \sum_{c \in \mathcal{C}(\mathbf{p}_{\text{NBV}})} \frac{V(c, \mathbf{p}_{\text{NBV}}) H(c)}{k(c)^{\beta}},
 \end{equation}
 
 \noindent where $k(c)$ denotes the rank (considering path length as metric) of the shortest path hypotheses traversing cell $c$ and $\beta$ is a parameter that can be used to give more weight to cells lying along the shortest path hypothesis. 
 
 The distance term is computed as 
  \begin{equation}
     J_{d}(\mathbf{p}_{\text{NBV}}) = \sum_{c \in \mathcal{C}'(\mathbf{p}_{\text{NBV}})} \frac{\max (0, d(c) - d(c, \mathbf{p}_{\text{NBV}}))}{k(c)^{\beta}},
 \end{equation}
 where $\mathcal{C}'(\mathbf{p}_{\text{NBV}})$ denotes the cells having $V(c, \mathbf{p}_{\text{NBV}}) > \gamma$, i.e. at least some probability of being visible from $\mathbf{p}_{\text{NBV}}$, $d(c)$ is the Euclidean distance between the robot's current pose and the centroid of $c$, and $d(c, \mathbf{p}_{\text{NBV}})$ is the Euclidean distance between the candidate NBV pose and cell $c$.

After having computed the objective function value for all the candidate NBV poses in $\mathcal{P}$, we sort them by decreasing value and try to relink them to the original path hypotheses via short trajectory segments obtained by growing a new RRT from the candidate NBV pose. We attempt to relink at most $N_{\text{REL}}$ NBV poses to the original hypotheses by growing RRTs from them. If this process is not successful, the planner again proceeds to Stage 3 and tries to follow the shortest hypothesis. In case of success, the planner can move to Stage 2. Fig.~\ref{fig:qual} shows the result of the procedure just described during one of our simulations. The NBV pose is shown in red (with the corresponding common trajectory segment) and the two path hypotheses in blue. Current cells in $\mathcal{N}$ are shown in blue, while pink cells denote unknown ground cells in $\mathcal{U}$.

\subsection{Stage 2: Navigation to NBV,  Hypothesis Commitment}

During this stage, the robot uses its local controller to move to the candidate NBV pose, while simultaneously updating the map. Fig.~\ref{fig:example}(a) shows the map status at this point (continuing the example of Fig.~\ref{fig:qual}). Note that measurements received during this stage are not explicitly accounted for in the choice of the NBV pose; this makes planning at the previous stage much faster, enabling a real-time implementation, but might still contribute to the overall uncertainty reduction. If there exists a path $\pi \in \mathcal{H}$ such that $|c(\mathbf{p}_i) \cap \mathcal{F}| = |c(\mathbf{p}_i)|$ for all poses $\mathbf{p}_i$ belonging to $\pi$ (namely, all corresponding cells are safe to traverse) then $\pi$ is safe and the robot can commit to following it until the goal is reached. Otherwise, the hypotheses in $\mathcal{H}$ containing at least one pose $\mathbf{p}_i$ such that $|c(\mathbf{p}_i) \cap \mathcal{O}| \neq \emptyset$ are removed from $\mathcal{H}$. If no hypotheses are left in $\mathcal{H}$, an additional check is made to ensure that the goal region $X_{\text{goal}}$ is actually reachable. If it is not, planning is aborted. Otherwise, the planner computes a single new path and proceeds to Stage 3. If at least one hypothesis is left in $\mathcal{H}$, the planner computes one or two new hypotheses to try to improve the previous ones; different heuristics could be used to quantify  improvement. In our current implementation, old hypotheses are replaced only if the new ones traverse less unknown and uncertain cells. After this, if a single hypothesis is left, the planner commits to it; otherwise, the planner commits to the shortest one. Fig.~\ref{fig:example}(b)  shows the commitment to a new hypothesis at the end of Stage 2.

\subsection{Stage 3: Hypothesis Following and Replanning} During this stage, the planner behaves as a typical online planner. At each replanning cycle, a new shortest path is computed with A* under the traditional assumption that unknown cells are actually free. This is done until the goal is reached, as in Fig.~\ref{fig:example}(c), or proven unreachable.



 

%% file: content/experiments.tex
\section{SIMULATIONS}
\label{sec:simulations}

\begin{figure}[t!]
    \centering
    \includegraphics[width=\columnwidth]{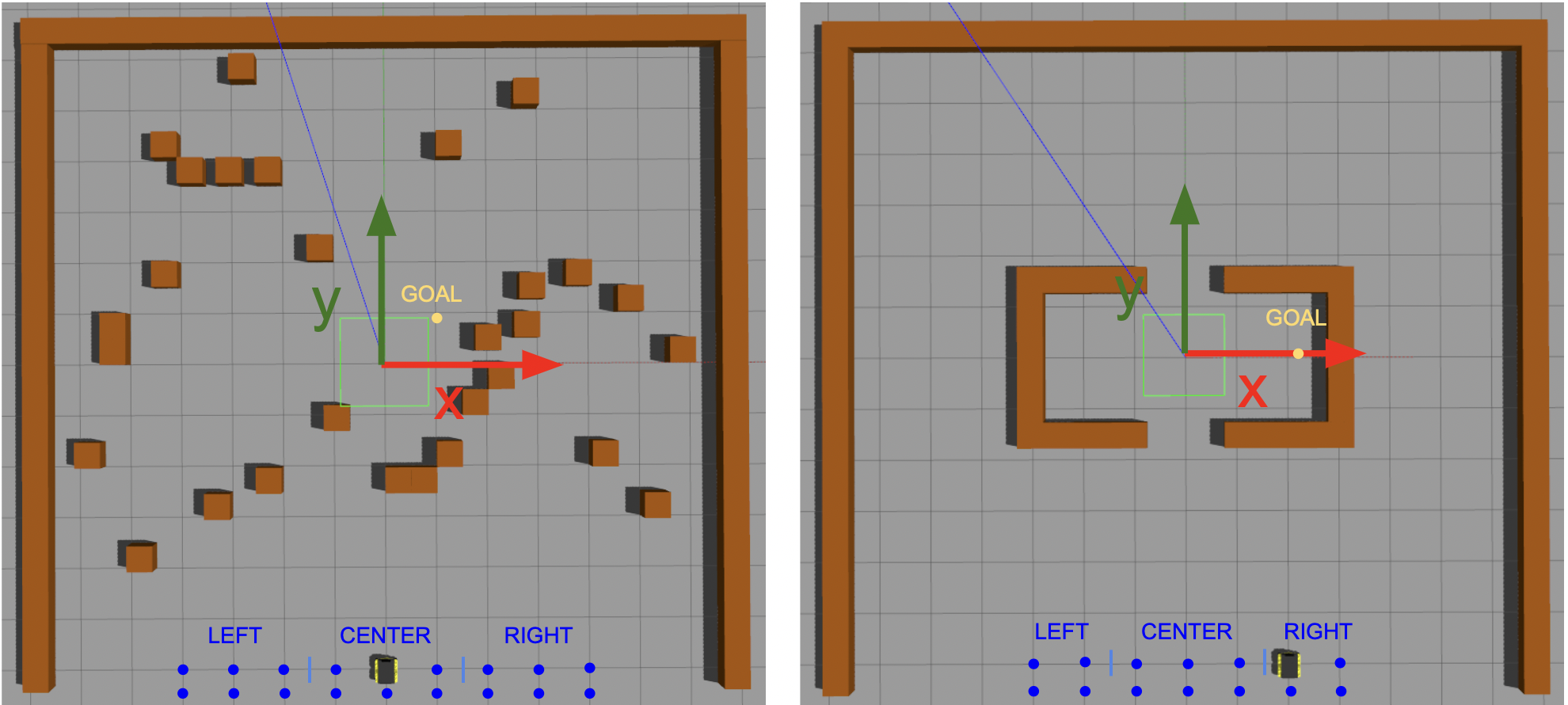}
    \caption{Simulation environments (1m$\times$1m cells). Left: Open; Right: Room. Jackal and blue dots denote start positions, while the goal position is in yellow. Best viewed when zoomed in.}
    \label{fig:environments}
\end{figure}

\begin{figure}[t!]
    \centering
    \includegraphics[width=0.8\columnwidth]{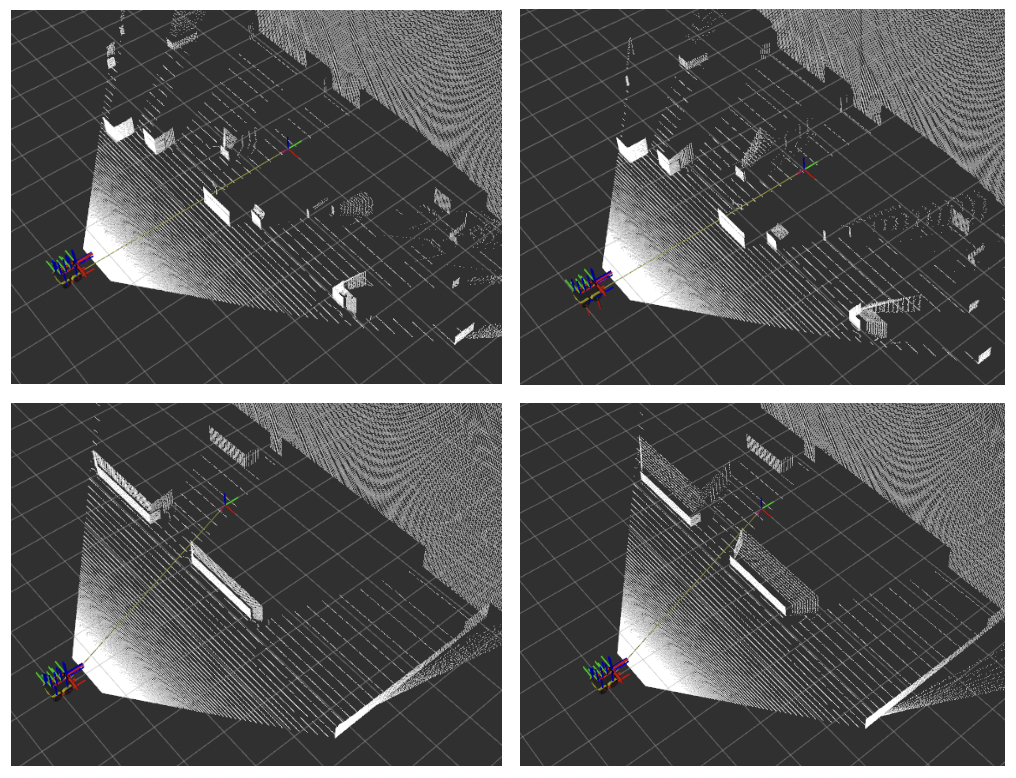}
    \caption{Simulated noise obtained from the robot's positions in Fig.~\ref{fig:environments}. Top: Open environment; Bottom: Room environment. Left: low noise; Right: high noise. The ``wall of points'' in the background lies beyond the mapping range and helps Octomap clearing free cells lying inside the robot's FOV.}
    \label{fig:noise}
\end{figure}

\subsection{Noise Simulation}
In order to study the impact of noise on planning without focusing on a particular depth camera sensor, we conduct an extensive set of simulations in Gazebo~\cite{gazebo} in two representative environments shown in Fig.~\ref{fig:environments}. The ``Open'' environment contains several small obstacles scattered onto the floor, and resulting on a dead-end on the right when facing upwards; the ``Room'' environment contains a symmetric rectangular structure with two entrances placed at the top and bottom. We simulate a Jackal Robot equipped with a camera providing depth measurements (672x376, 107 degree FOV). In order to study the impact of different noise levels on the planners' performance, the following procedure is used to corrupt the ground truth depth provided by Gazebo with the typical noise experienced by depth cameras, which is concentrated around the objects' edges and increases with the square of the distance~\cite{wang2019pseudo}. First, we use the Canny edge detector to detect edge pixels in the current depth image, corresponding to abrupt changes in depth. To mitigate imperfections in the detections, pixels on the horizon are ignored, and pixels that fall onto the background due to aliasing are adjusted to be relocated onto their adjacent obstacles. Then, for each edge pixel $p_e$ lying beyond a ``perfect sensing'' distance $s=3$ meters from the camera, we consider a square window of fixed size $w=10$ pixels, and update the current depth $i(p_w)$ (measured in meters) of each pixel $p_w$ in that window with Euclidean distance $d(p_e,p_w) \leq w$ as:

\begin{small}
\begin{equation*}
i(p_w) = \max\bigg{[}i(p_w), \rho\cdot i(p_w) \cdot \bigg{(}w - d(p_e, p_w)\bigg{)} \cdot d(s, i(p_w))^2 + \sigma \bigg{]},
\end{equation*}
\end{small}

\noindent where $\sigma \sim \mathcal{N}(a \cdot d(s, i(p_w)), b \cdot d(s, i(p_w))^2)$. The simulation parameters are hence $\rho$, $a$, and $b$. 
Also note that we take the maximum between the current and new value since the same pixel might appear in multiple edge pixel windows. Fig.~\ref{fig:noise} shows the effect of two different choices of parameter combinations on a point cloud obtained in the two environments: $\rho=0.01$, $a=0.05$, $b=0.002$ (low), and $\rho=0.05$, $a=0.125$, $b=0.005$ (high). 

\begin{table*}[]
\scriptsize
\centering
\subfloat[Open Environment]{
\begin{tabular}{|l|c|c|c|c|}
\hline
\multicolumn{5}{|l|}{\bf Low Noise ($\rho=0.01, a=0.05,b=0.002$)}                                                         \\ \hline
Start                    & \multicolumn{1}{l|}{Planner} & S / U / F & Dist. $\pm$ 95\% C.I. & Diff. \\ \hline
\multirow{3}{*}{Left} & NBV                          & 119 / 0 / 1       & 9.47 $\pm$ 1.30   & N.A.   \\ \cline{2-5} 
                         & B3.5                         & 120 / 0 / 0       & 8.06 $\pm$ 0.17   & -15\%   \\ \cline{2-5} 
                         & B7                        & 120 / 0 / 0       & 8.14 $\pm$ 0.21   & -14\%   \\ \hline
\multirow{3}{*}{Center} & NBV                          & 109 / 2 / 9       & 12.28 $\pm$ 1.82   & N.A.   \\ \cline{2-5} 
                         & B3.5                         & 109 / 0 / 11       & 10.38 $\pm$ 1.81   & -15\%   \\ \cline{2-5} 
                         & B7                         & 117 / 0 / 3       & 8.41 $\pm$ 0.87   & -32\%   \\ \hline
\multirow{3}{*}{Right} & NBV                          & 116 / 0 / 4       & 12.30 $\pm$ 0.70   & N.A.   \\ \cline{2-5} 
                         & B3.5                         & 112 / 0 / 8       & 15.35 $\pm$ 0.83   & \textcolor{ForestGreen}{\bf +25\%}   \\ \cline{2-5} 
                         & B7                         & 118 / 0 / 2     & 9.42 $\pm$ 0.74   & -23\%   \\ \hline
\multicolumn{5}{|l|}{\bf High Noise ($\rho=0.05, a=0.125, b=0.005$)} \\ \hline
\multirow{3}{*}{Left} & NBV                          & 118 / 0 / 2       & 8.92 $\pm$ 1.10   & N.A.   \\ \cline{2-5} 
                         & B3.5                         & 120 / 0 / 0       & 8.10 $\pm$ 0.19   & -9\%   \\ \cline{2-5} 
                         & B7                        & 120 / 0 / 0       & 8.25 $\pm$ 0.48   & -8\%  \\ \hline
\multirow{3}{*}{Center} & NBV                          & 116 / 1 / 4       & 10.02 $\pm$ 1.36   & N.A.   \\ \cline{2-5} 
                         & B3.5                         & 116 / 0 / 4       & 10.10 $\pm$ 1.77   & $<$ +1\%   \\ \cline{2-5} 
                         & B7                         & 112 / 0 / 8       & 8.78 $\pm$ 1.25   & -12\%   \\ \hline
\multirow{3}{*}{Right} & NBV                          & 118 / 1 / 1       & 12.42 $\pm$ 0.97   & N.A.   \\ \cline{2-5} 
                         & B3.5                         & 107 / 0 / 13       & 15.47 $\pm$ 0.50   & \textcolor{ForestGreen}{\bf +25\%}    \\ \cline{2-5} 
                         & B7                         & 44 / \textcolor{red}{\bf 74} / 0       & 9.73 $\pm$ 2.15   & -22\%   \\ \hline
\end{tabular}}
\qquad\qquad
\subfloat[Room Environment]{%
\begin{tabular}{|l|c|c|c|c|}
\hline
\multicolumn{5}{|l|}{\bf Low Noise ($\rho=0.01, a=0.05,b=0.002$)}                                                         \\ \hline
Start                    & \multicolumn{1}{l|}{Planner} & S / U / F & Dist. $\pm$ 95\% C.I. & Diff. \\ \hline
\multirow{3}{*}{Left} & NBV                          & 80 / 0 / 0       & 8.29 $\pm$  0.67  & N.A.   \\ \cline{2-5} 
                         & B3.5                         &  80 / 0 / 0       & 7.98 $\pm$ 0.16   & -4\%   \\ \cline{2-5} 
                         & B7                         &  80 / 0 / 0       & 7.99 $\pm$ 0.13   & -4\%   \\ \hline
\multirow{3}{*}{Center} & NBV                          & 115 / 0 / 5       & 8.38 $\pm$ 0.63   & N.A.   \\ \cline{2-5} 
                         & B3.5                         & 120 / 0 / 0       & 8.54 $\pm$ 0.71   & +2\%   \\ \cline{2-5} 
                         & B7                         & 120 / 0 / 0       & 7.04 $\pm$ 0.14   & -16\%   \\ \hline
\multirow{3}{*}{Right} & NBV                          & 76 / 0 / 4       & 11.66 $\pm$ 1.25  & N.A.   \\ \cline{2-5} 
                         & B3.5                         &  79 / 0 / 1       & 11.97 $\pm$ 1.05  & +3\%   \\ \cline{2-5} 
                         & B7                         & 79 / 0 / 1       & 10.21 $\pm$ 0.44  & -12\%   \\ \hline
\multicolumn{5}{|l|}{\bf High Noise ($\rho=0.05, a=0.125, b=0.005$)} \\ \hline
\multirow{3}{*}{Left} & NBV                          & 80 / 0 / 0       & 8.29 $\pm$ 0.46   & N.A.   \\ \cline{2-5} 
                         & B3.5                         & 80 / 0 / 0       & 7.96 $\pm$ 0.16   & -4\%   \\ \cline{2-5} 
                         & B7                        & 1 / \textcolor{red}{\bf 79} / 0      & 7.51   & -9\%   \\ \hline
\multirow{3}{*}{Center} & NBV                          & 114 / 0 / 6       & 8.30 $\pm$ 0.39   & N.A.   \\ \cline{2-5} 
                         & B3.5                         & 120 / 0 / 0       & 8.52 $\pm$ 0.66   & +3\%   \\ \cline{2-5} 
                         & B7                         & 62 / \textcolor{red}{\bf 58} / 0       & 7.16 $\pm$ 0.22   & -14\%   \\ \hline
\multirow{3}{*}{Right} & NBV                          & 79 / 0 / 1       & 11.23 $\pm$ 0.94   & N.A.   \\ \cline{2-5} 
                         & B3.5                         & 80 / 0 / 0       & 11.39 $\pm$ 0.81  & +1\%   \\ \cline{2-5} 
                         & B7                         & 39 / \textcolor{red}{\bf 40} / 1       & 10.12 $\pm$ 0.70   & -10\%   \\ \hline
\end{tabular}}
\caption{Simulations results. A high number of runs with goal declared unreachable is highlighted in red. In green, a large positive difference ($>$ 20\%) in avg. traveled distance between NBV planner and short-range baseline. }
\end{table*}

\subsection{Mapping}

We use the Octomap mapping framework~\cite{hornung13auro} for map building. The original implementation uses a beam-based inverse sensor model for processing the measurements. Although this is not specifically designed for depth cameras (for which more sophisticated techniques could be used, see~\cite{stereomodel}), it can still provide a good approximation. The mapping range is set to 7 meters, and we assume that measurements within 3.5 meters can be considered reliable. As a heuristic to cope with less accurate measurements after 3.5 meters,  we introduce a smaller upper clamping threshold for the occupancy probability of the Octomap leaves receiving measurements lying above the ground and between 3.5 and 7 meters from the camera (specifically, 0.7) compared to the one used between 0 and 3.5m (0.85). The reader is referred to the Octomap paper~\cite{hornung13auro} for more details on the role of clamping thresholds. The octomap resolution is set to 0.25m, and measurements are integrated into the map at 2Hz after a downsampling of the pointcloud with a 0.25m resolution. Probability of hit and miss for the beam-based model are set to 0.7 and 0.2, respectively.

\subsection{NBV Planner Implementation}
The NBV planner is implemented in C++ as a global planner plugin of the ROS \texttt{move\_base}\footnote{\url{http://wiki.ros.org/move_base}} package, with replanning happening at 1 Hz. In all simulations, hypotheses planning and replanning with A* is instantaneous, while the NBV computation is $\approx$130 ms on average\footnote{Our current implementation is sequential, but this operation could be easily parallelized.} on a computer equipped with an Intel i7-7740X  CPU. The default \texttt{move\_base} local planner is used to track the computed trajectories. After some tuning, the following parameters are selected for the NBV function $J()$: $\alpha=0.5$, $\beta=2$, $\gamma=0.05$. The remaining planner parameters are set as follows: $|\mathcal{P}|=200$, $C_{\mathcal{N}}=2$, $D_{\text{HYP}}=4$, $R_{\text{RRT}}=4$ meters, $N_{\text{REL}}=7$. The $p_l$ and $p_h$ uncertainty thresholds are set to 0.18 and 0.75, respectively.

\subsection{Baselines}

The traditional planning approach described at the end of Section~\ref{sec:problem} is used as a baseline, where a single threshold probability equal to 0.3 is used to determine if a cell with unclear occupancy is free or occupied. We consider two versions of this planner: one where the mapping range is limited to 3.5 meters, and one where the mapping range is equal to the one used for the NBV planner (7 meters). In the former case, measurements are almost always accurate, but the short mapping range might result in myopic behaviors. In the latter case, cells are often treated as obstacles even when they are actually free.

\subsection{Performance Metrics}

We consider the traveled distance as main performance metric. We also examine how many times the goal is reached or declared unreachable, and how many times the robot fails to reach the goal within 2 minutes. The latter case is typically caused by the local planner getting stuck in proximity of an obstacle ---an event not directly caused by the global planners due to the absence of noise at close range.

\subsection{Open Environment - Results}

We place the goal at $(1, 1)$ and consider start positions resulting from combining $x$ coordinates in $\{0,\pm 1, \pm 2, \pm 3, \pm 4\}$ and $y$ in $\{-6, -5.5\}$. Start positions are divided into three groups depending on their $x$ coordinate: center ($0, \pm 1$), left ($-2, -3, -4$), and right ($2, 3, 4$). The initial orientation is kept fixed to $90$ degrees. We execute 20 runs for each start position. Results are reported in Table 1. Here, the ``S / U / F'' column denotes the number of successes, failures due to goal unreachable, and local planner failures. The average distance is reported with 95\% confidence intervals, and the ``Diff.'' column reports the difference in percentage between the average distance traveled by the NBV planner and the baselines. 

The simulation results show a number of interesting characteristics. First, the long-range baseline (B7) outperforms the short-range baseline (B3.5) and the NBV planner in all groups of instances in terms of traveled distance. However, it is not able to reach the goal in a significant number of cases when starting from the right, which makes it not very competitive against the other two planners. For the NBV and short-range baselines, we observe slightly better performance of the latter when starting from the center-left, regardless of the noise level. This happens for two reasons. First, from those regions, there is almost always a straight path to the goal (except for $x=1$). Second, the NBV planner might often make slight detours to the reach the chosen NBV due to the local planner trying to closely follow the first trajectory segment (an RRT branch). We believe that a smarter implementation of the relinking phase, in which the traveled distance is optimized (e.g. with RRT$^*$) could easily mitigate this second small issue.  When starting from the right, however, the NBV planner is the clear winner, with average improvements of 25\% for both high and low noise: the myopic behavior of the short-range baseline always causes the robot to travel almost all the way up to the dead-end before realizing that the goal cannot be reached from the right. 

\subsection{Room Environment - Results}

We place the goal at $(2.2, 0)$ and consider start positions resulting from combining $x$ coordinates in $\{0,\pm 1, \pm 2, \pm 3\}$ and $y$ in $\{-6, -5.5\}$ (20 total). Start positions are divided into three groups depending on their $x$ coordinate: center ($0, \pm 1$), left ($-2, -3$), and right ($2, 3$). The initial orientation is again kept fixed to $90$ degrees. We execute 20 runs for each start position. The results are reported in Table 2. Again, the long-range baseline often declares the goal to be unreachable in presence of high noise. This makes it a not very compelling choice regardless of the environment. Comparing the NBV planner with the short-range baseline, we now observe a more uniform performance between the two compared to the Open environment. Even when starting from the center-right, the NBV is only able to offer a very small improvement (1-3\%). This mainly happens because the shape of the ``room'' is such that, when the baseline starts taking the longer path to the right, it immediately perceives the wall to the left and, depending on the number of obstacle cells that are perceived, it may still quickly realize that turning back would make the robot travel less distance.


\section{REAL WORLD VALIDATION}

\begin{figure}[t!]
  \centering
 \includegraphics[width=\columnwidth]{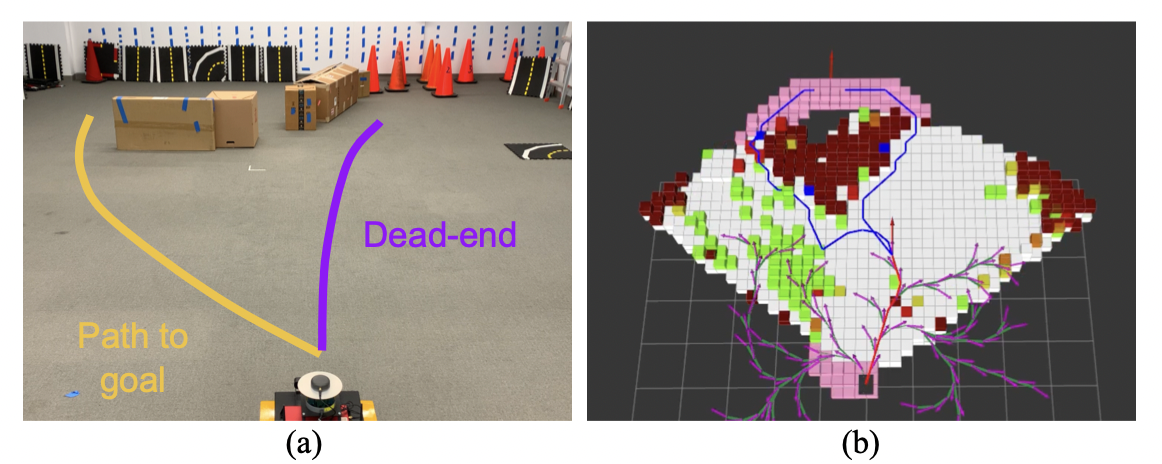}
\caption{Real world experiment. (a): Environment; (b): Initial Octomap map, with path hypotheses and NBV.}
\label{fig:real_world}
 \vspace*{-3mm}
\end{figure}

We validate the proposed NBV planner with a Jackal robot equipped with a ZED 2 stereo camera.\footnote{\url{https://www.stereolabs.com/zed/}} The environment used for the experiment is shown in Fig.~\ref{fig:real_world} (a). The ZED 2 software is used to provide the Jackal with a pose estimate. The goal of this experiment is to study whether the NBV and baseline planners show a behavior similar to that observed in simulation in the Open environment, as both environments contain a dead-end on the right and a path to the goal on the left. The noise experienced in the real world resulted in grid maps visually similar to the simulated ones. However, we had to slightly reduce the mapping range to 6.5 meters in order to make the goal eventually reachable due to the presence of more depth noise than expected when observing the goal from the right.

Across 10 runs, the NBV planner reaches the goal 7 times, traveling an average distance of 11.35m (std. dev. 0.94m). The goal is decleared unreachable 2 times, and we register one failure due the robot getting stuck in proximity of an obstacle. The typical behavior for the 7 successes is shown in Fig.~\ref{fig:real_world}(b) and is consistent to what observed in simulation: an informative view is taken on the right, and allows the robot to realize the presence of the dead-end. Consistently with the simulations, the long-range baseline declares the long unreachable  more often (5 times), but travels an average distance of 8.65m when the goal is reached (std. dev. 0.28m). When planning with the short-range baseline the goal is reached 8 times (1 goal unreachable, 1 failure). However, the robot is often tricked into driving into the dead-end before realizing that the path is blocked. This makes the robot travel more distance on average (13.52m, std. dev. 3.69m) compared to the NBV planner.

%% file: content/discussion.tex
\section{DISCUSSION}

In this paper, we studied the impact of occupancy map uncertainty on global path planning, and proposed a novel planning approach to directly take into account map uncertainty. Overall, the inclusion of probabilistic reasoning about the map and next best view in the planner, along with more than one plan hypotheses, led to more reliable plans with strong performance compared to short- and long-range baseline implementations whenever one or more obstacles would preclude the existence of a straight path to the goal. At the same time, our results showed that, in those cases where the path planning problem is inherently simple, ``overthinking'' the path to take might be counterproductive, and a more myopic baseline planner could work just as well. Data-driven techniques, and especially deep learning, might hold the key to discriminate between these different types of instances. Other interesting research directions are related to the study of this problem with more sophisticated sensor models and/or environment representations based on point clouds.

%% file: main.bbl
\begin{thebibliography}{10}
\providecommand{\url}[1]{#1}
\csname url@samestyle\endcsname
\providecommand{\newblock}{\relax}
\providecommand{\bibinfo}[2]{#2}
\providecommand{\BIBentrySTDinterwordspacing}{\spaceskip=0pt\relax}
\providecommand{\BIBentryALTinterwordstretchfactor}{4}
\providecommand{\BIBentryALTinterwordspacing}{\spaceskip=\fontdimen2\font plus
\BIBentryALTinterwordstretchfactor\fontdimen3\font minus
  \fontdimen4\font\relax}
\providecommand{\BIBforeignlanguage}[2]{{%
\expandafter\ifx\csname l@#1\endcsname\relax
\typeout{** WARNING: IEEEtran.bst: No hyphenation pattern has been}%
\typeout{** loaded for the language `#1'. Using the pattern for}%
\typeout{** the default language instead.}%
\else
\language=\csname l@#1\endcsname
\fi
#2}}
\providecommand{\BIBdecl}{\relax}
\BIBdecl

\bibitem{wang2019pseudo}
Y.~Wang, W.-L. Chao, D.~Garg, B.~Hariharan, M.~Campbell, and K.~Q. Weinberger,
  ``Pseudo-lidar from visual depth estimation: Bridging the gap in 3d object
  detection for autonomous driving,'' in \emph{Proc. CVPR}, 2019, pp.
  8445--8453.

\bibitem{hornung13auro}
\BIBentryALTinterwordspacing
A.~Hornung, K.~M. Wurm, M.~Bennewitz, C.~Stachniss, and W.~Burgard,
  ``{OctoMap}: An efficient probabilistic {3D} mapping framework based on
  octrees,'' \emph{Autonomous Robots}, 2013, software available at
  \url{http://octomap.github.com}. [Online]. Available:
  \url{http://octomap.github.com}
\BIBentrySTDinterwordspacing

\bibitem{ryll2019efficient}
M.~Ryll, J.~Ware, J.~Carter, and N.~Roy, ``Efficient trajectory planning for
  high speed flight in unknown environments,'' in \emph{Proc. ICRA}, 2019, pp.
  732--738.

\bibitem{tordesillas2020faster}
J.~Tordesillas, B.~T. Lopez, M.~Everett, and J.~P. How, ``Faster: Fast and safe
  trajectory planner for flights in unknown environments,'' \emph{arXiv
  preprint arXiv:2001.04420}, 2020.

\bibitem{hart1968formal}
P.~E. Hart, N.~J. Nilsson, and B.~Raphael, ``A formal basis for the heuristic
  determination of minimum cost paths,'' \emph{IEEE transactions on Systems
  Science and Cybernetics}, vol.~4, no.~2, pp. 100--107, 1968.

\bibitem{stentz1994}
A.~Stentz, ``Optimal and efficient path planning for partially known
  environments,'' in \emph{Proc. ICRA}, 1994.

\bibitem{stentz1995}
------, ``The focussed {D}* algorithm for real-time replanning,'' in
  \emph{Proc. IJCAI}, 1995.

\bibitem{koenig2002d}
S.~Koenig and M.~Likhachev, ``D* lite,'' in \emph{Proc. AAAI}, 2002, pp.
  476--483.

\bibitem{rrt}
S.~LaValle, ``Rapidly-exploring random trees: A new tool for path planning,''
  Tech. Rep., 1998.

\bibitem{kavraki1996probabilistic}
L.~E. Kavraki, P.~Svestka, J.-C. Latombe, and M.~H. Overmars, ``Probabilistic
  roadmaps for path planning in high-dimensional configuration spaces,''
  \emph{IEEE transactions on Robotics and Automation}, vol.~12, no.~4, pp.
  566--580, 1996.

\bibitem{dolgov2008practical}
D.~Dolgov, S.~Thrun, M.~Montemerlo, and J.~Diebel, ``Practical search
  techniques in path planning for autonomous driving,'' in \emph{Proc. STAIR},
  2008.

\bibitem{harabor2011online}
D.~D. Harabor, A.~Grastien \emph{et~al.}, ``Online graph pruning for
  pathfinding on grid maps.'' in \emph{Proc. AAAI}, 2011, pp. 1114--1119.

\bibitem{daniel2010theta}
K.~Daniel, A.~Nash, S.~Koenig, and A.~Felner, ``Theta*: Any-angle path planning
  on grids,'' \emph{Journal of Artificial Intelligence Research}, vol.~39, pp.
  533--579, 2010.

\bibitem{stein2018learning}
G.~J. Stein, C.~Bradley, and N.~Roy, ``Learning over subgoals for efficient
  navigation of structured, unknown environments,'' in \emph{Proc. CORL}, 2018,
  pp. 213--222.

\bibitem{yutao2}
Y.~Han, J.~Banfi, and M.~Campbell, ``Planning paths through unknown space by
  imagining what lies therein,'' in \emph{Proc. CORL}, 2020, to appear.

\bibitem{dshppc2020}
Y.~Han, H.~Lin, J.~Banfi, M.~Campbell, and K.~Bala, ``{DeepSemanticHPPC}:
  Hypothesis-based planning over uncertain semantic point clouds,'' in
  \emph{Proc. ICRA}, 2020, pp. 4252--4258.

\bibitem{hardy}
J.~Hardy and M.~Campbell, ``Contingency planning over probabilistic obstacle
  predictions for autonomous road vehicles,'' \emph{IEEE Trans Robot}, vol.~29,
  no.~4, pp. 913--929, 2013.

\bibitem{daudelin2017adaptable}
J.~Daudelin and M.~Campbell, ``An adaptable, probabilistic, next-best view
  algorithm for reconstruction of unknown 3-d objects,'' \emph{IEEE Robotics
  and Automation Letters}, vol.~2, no.~3, pp. 1540--1547, 2017.

\bibitem{gazebo}
{Open Source Robotics Foundation}, ``Gazebo. robot simulation made easy,''
  \url{http://gazebosim.org/}, 2014.

\bibitem{stereomodel}
F.~{Andert}, ``Drawing stereo disparity images into occupancy grids:
  Measurement model and fast implementation,'' in \emph{Proc. IROS}, 2009, pp.
  5191--5197.

\end{thebibliography}
